\documentclass{article}					

\PassOptionsToPackage{numbers, sort&compress}{natbib}

	\usepackage[preprint]{neurips_2019}

\usepackage[utf8]{inputenc}							
\usepackage[T1]{fontenc}							
\usepackage{hyperref}								
\usepackage{url}									
\usepackage{booktabs}								
\usepackage{amsfonts}								
\usepackage{amsmath}								
\usepackage{nicefrac}								
\usepackage{microtype}								
\usepackage[]{graphicx}								
\usepackage{subcaption}								
\usepackage[ruled,longend]{algorithm2e}				

\title{Improving Neural Networks by Adopting Amplifying and Attenuating Neurons}

\author{
	Seongmun Jung\\
	Department of Aerospace Engineering \\
	KAIST, Daejeon, Republic of Korea \\
	\texttt{mmx1517@kaist.ac.kr} \\
	\And
	Oh Joon Kwon \\
	Department of Aerospace Engineering \\
	KAIST, Daejeon, Republic of Korea \\
	\texttt{ojkwon@kaist.ac.kr} \\
}

\begin{document}

\maketitle

\begin{abstract}
In the present study, an amplifying neuron and attenuating neuron, which can be easily implemented into neural networks without any significant additional computational effort, are proposed.
The activated output value is squared for the amplifying neuron, while the value becomes its reciprocal for the attenuating one.
Theoretically, the order of neural networks increases when the amplifying neuron is placed in the hidden layer.
The performance assessments of neural networks were conducted to verify that the amplifying and attenuating neurons enhance the performance of neural networks.
From the numerical experiments, it was revealed that the neural networks that contain the amplifying and attenuating neurons yield more accurate results, compared to those without them.
\end{abstract}

\section{Introduction}

Artificial neural networks with a sufficient number of neurons are universal approximators \citep{hornik1989multilayer};
therefore, properly trained networks with a sufficient number of hidden neurons and hidden layers are able to approximate any unknown functions.
Over the past few decades, the studies on the weights initialization \citep{glorot2010understanding, he2015delving} and activation functions \citep{nair2010rectified, clevert2015fast, he2015delving} were conducted, and now it is possible to train the neural networks having multiple hidden layers.
Hence, neural networks with multiple hidden layers have been successfully adopted to various areas, such as image classification \citep{he2016deep}, natural language processing \citep{socher2013recursive}, and numerical model improvement \citep{singh2017machine}.

Since neural networks are theoretically able to approximate any functions within small error ranges as its size increases \citep{he2016deep}, large neural networks are preferred for handling complicated tasks.
However, large neural networks also suffer from the gradient vanishing/exploding and degradation problem \citep{srivastava2015highway}.
In addition, as the network size gets larger, the required computational effort increases.
Recently, Residual Network \citep{he2016deep} and Batch Normalization \citep{ioffe2015batch} enable to train large neural networks.
Nonetheless, additional computations are required, and the size of neural networks is still limited depending on the computational power available.
Therefore, it is crucial to enhance the efficiency and performance of neural network models under the constraint of a specified network size.

From the early neural network models \citep{rosenblatt1958perceptron, fukushima1975cognitron, fukushima1980neocognitron}, which were developed by imitating biological nervous systems, several studies have been conducted to improve the neural network model.
For instance, instead of squashing functions, which are similar to the “all-or-none” activation process  \citep{mcculloch1943logical} of the real neuron, the Rectified Linear Unit (ReLU) function \citep{nair2010rectified} is widely used as an activation function for neural networks.
It was also found that neural networks yield improved results when the squared value of the input component is augmented into the input layer \citep{flake1998square}, and when Batch Normalization method \citep{ioffe2015batch} is adopted.

In the present study, an amplifying neuron and attenuating neuron are proposed in order to improve the neural network model.
The amplifying neuron amplifies the activated value of the neuron, and the attenuating neuron attenuates the activated value.
Those amplifying and attenuating neurons can be readily implemented into the neural network model by applying secondary activation functions.
Theoretically, neural networks are possible to perform the multiplication and approximated division of the inputs with the amplifying and attenuating neurons.
In addition, the amplifying neuron makes the order of neural networks higher.
Therefore, the neural network that contains the amplifying neuron is expected to address the tasks that are more complicated.
Such characteristics make it possible to use smaller neural networks for identical tasks, and to obtain smaller error ranges for the neural networks of an identical size.

\section{Neural networks}

\begin{figure}
	\centering
	\begin{subfigure}{0.45\linewidth}
		\centering
		\includegraphics[width=\linewidth]{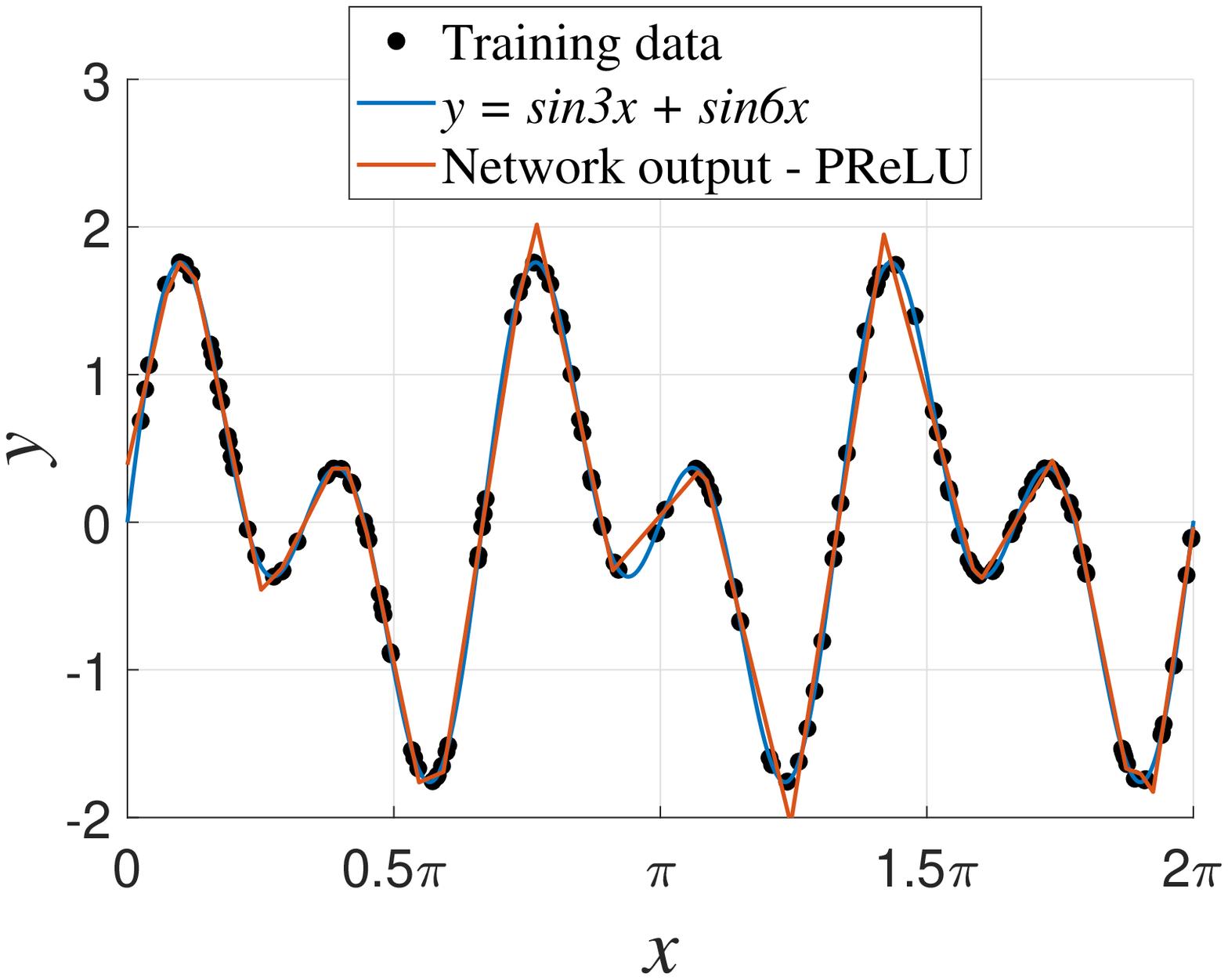}
		\caption{The training data set, exact solution, and output of the network}
	\end{subfigure}
	\begin{subfigure}{0.45\linewidth}
		\centering
		\includegraphics[width=\linewidth]{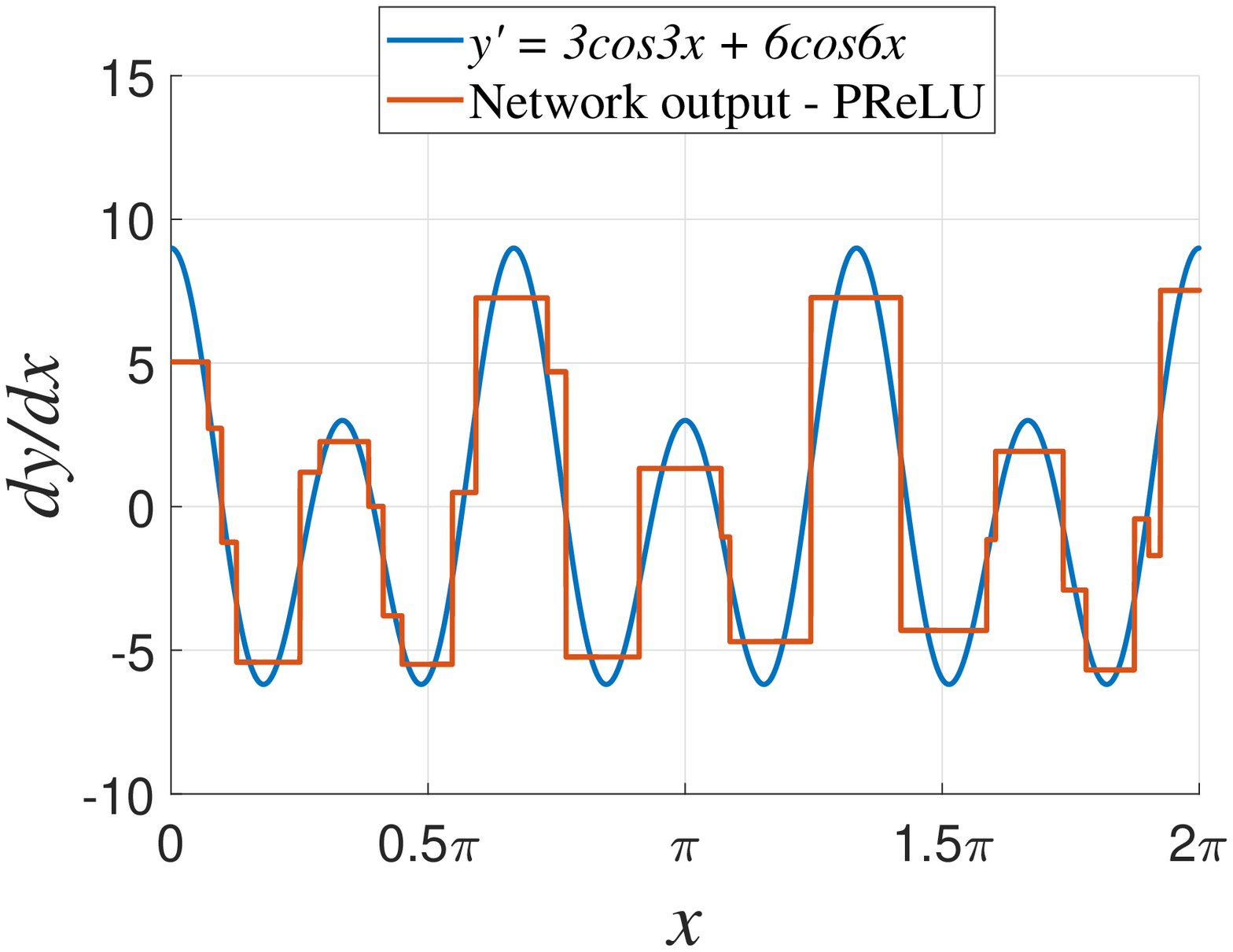}
		\caption{The derivative of the exact solution and network output}
	\end{subfigure}
	\caption{The exact solution and results of the neural network that consists of four hidden layers and seven hidden neurons per hidden layer.}
	\label{fig:sinosoidal}
\end{figure}
Neural networks are often referred as black boxes, since it is hard to figure out how they work.
Especially, the difference between neural networks and biological nervous systems makes it difficult to apply biological intuitions to the neural networks model.
For instance, the Heaviside step function tends to be avoided as an activation function, in spite of its similarity to the “all-or-none” law, since its derivative is zero throughout the whole domain.
Therefore, the Heaviside function is not suitable for the error back-propagation \citep{rumelhart1988learning} learning process.
Nonetheless, the neural network model is fundamentally an algebraic model that consists of non-linear activations and matrix multiplication operations.
Hence, it is possible to conduct an algebraic and numerical analysis of the neural network model.
\begin{equation} \label{eq:neural_network}
	\boldsymbol{y}^{k+1}=\boldsymbol{F}(\boldsymbol{W}^k\boldsymbol{y}^k)
\end{equation}
In the neural network model, the output vector of each layer is calculated by using the output vector of the previous layer as in Eq. (\ref{eq:neural_network}).
Therefore, the range of the previous layer becomes the domain of the following layer, and this process can be expressed via the function composition.
Since the composite function of two linear functions is also a linear function, the output of the neural network consists of linear elements, as shown in figure \ref{fig:sinosoidal}, when the piecewise linear activation functions without any curvature (e.g., the ReLU and LReLU) are used. 
Furthermore, the derivative of the neural network output is composed of discontinuous constant elements, and it can be considered as the linear combination of the Heaviside step functions with different biases.
Such characteristics imply that the neural network with the activation function of the Heaviside step function can be obtained by differentiating the neural network with piecewise linear activation functions.

It is not surprising that the neural network with the Heaviside step activation function is the derivatives of the network with the ReLU, as the Heaviside step function is the derivative of the ReLU.
Namely, the neural network model is the integrated model of the biological nervous system.
Therefore, the order of the neural network model is higher than the biological one, and less number of neurons are expected to be required for identical tasks.
Nonetheless, the neural network model is still needed to be improved.
In the case of one-dimensional input and output, neural networks yield similar results to the Euler method, which is generally avoided because of its accuracy.

One approach to the problem is to use the modified piecewise linear activation function having curvature, such as the ELU \citep{clevert2015fast} and softplus \citep{dugas2001incorporating}.
With those activation functions, the output derivative of the neural network becomes continuous;
hence, the performance of the neural network increases.
However, it is clear that the improvement of the neural network with those activation functions is limited, as they are highly similar to piecewise linear activation functions.
Another approach is to implement a high-order polynomial as the non-linear activation function of the network \citep{chen1993conventional, livni2014computational}.
Although such approach may increase the order of the neural network, the neural network can easily become divergent, especially when the degree of the polynomial activation function is high.
In addition, the improvement of the neural network performance is not guaranteed, because there exists no switching feature for the polynomial, and the derivative of the quadratic polynomial is not non-linear, but a linear function.

\section{Amplifying and attenuating neurons}

\subsection{Amplifying neurons}

As the output of the previous layer of the neural network becomes the input for the following layer, this process can be expressed via the function composition.
Therefore, the order of the neural network can be inferred by the function composition of the network activation function, provided that the activation function is continuous.
For instance, the neural network with the sigmoid activation function is a zeroth-order model, as the iterated function of the sigmoid function converges to a constant value function.
In a similar manner, the iterated functions of the ReLU, LReLU, PReLU, ELU, and softplus converge to the ReLU;
therefore, the neural network using those activation functions works as a first-order model.
\begin{equation} \label{eq:f_integ}
	\begin{cases}
		\int{f(x)dx} = \nicefrac{1}{2}\{f(x)\}^2      \qquad(f(x)=\text{ReLU})\\
		\int{f(x)dx} \approx \nicefrac{1}{2} \{f(x)\}^2\qquad(f(x)=\text{softplus})
	\end{cases}
\end{equation}
To improve the performance of the neural network, the order of the neural network model needs to be increased.
The order of the neural network increases when the high-order activation functions are used.
In the present study, an amplifying neuron that provides the high-order activation function is introduced.
The output of the amplifying neuron is determined to be the square of the activated value, since the exact integral of the ReLU is the square of the ReLU.
In addition, the square of the PReLU with $a$ is the exact integral of the PReLU with $a^{2}$.
Moreover, the integral of the softplus can be approximated by the squared softplus as in Eq. (\ref{eq:f_integ}).
Therefore, the amplifying neuron is an ordinary neuron with second-order activation functions when first-order activation functions are used.
Although the order of the sigmoid function does not increase by squaring, it is not an issue, since the softplus can be used to increase the order of the sigmoid function.
\begin{equation} \label{eq:amplification}
	\begin{cases}
		F(x)=G \circ H(x)\\
		G(x)=x^2
	\end{cases}
\end{equation}
The output of the amplifying neuron can be expressed using the function composition as in Eq. (\ref{eq:amplification}).
Hence, the amplifying neuron can easily be implemented into the neural network by applying the secondary activation function $G(x)$ with the original activation function $H(x)$.
The coefficient \nicefrac{1}{2} in Eq. (\ref{eq:f_integ}) is not considered in Eq. (\ref{eq:amplification}), as it can be considered by the weights between neurons.

\subsection{Attenuating neurons}

Although the amplifying neuron is the neuron that provides high-order activation functions, it can also be considered as the ordinary neuron that amplifies the input.
Here, an attenuating neuron, which performs an opposite operation compared to the amplifying neuron, is also introduced.
The main purpose of introducing this attenuating neuron is to provide multiplicative inverse value to the neural network, since it is inefficient to approximate such a value with high-order activation functions.
\begin{equation} \label{eq:attenuation}
	\begin{cases}
		F(x)=G \circ H(x)\\
		G(x)=\cfrac{x}{x^2+b} \approx \cfrac{1}{x}
	\end{cases}
\end{equation}
The attenuating neuron can be readily implemented into the neural network by using the secondary activation function $G(x)$ in a manner similar to the amplifying neuron implementation.
However, the reciprocal of $x$ cannot be directly used as the secondary activation function $G(x)$, since it contains a singularity where $x$ is zero.
To avoid this singularity of the reciprocal function, the modified reciprocal function without a singularity shown in Eq. (\ref{eq:attenuation}) is adopted as the secondary activation function $G(x)$ for the attenuating neuron with $b$ of one.

\subsection{Neural networks with amplifying and attenuating neurons}

\begin{algorithm}
	\SetAlCapSty{}					
	\DontPrintSemicolon 
	\For{$(i \gets 0;\; i<n_\text{layers};\; i \gets i + 1)$}
	{
		\For{$(j \gets 0;\; j<n_\text{neurons};\; j \gets j + 1)$}
		{
			Calculate $y(i, j)$\;
			Calculate $F'(i, j)$\;
			\;
			// Amplifying neurons\;
			\If{(neuron$(i,j)$ is the amplifying neuron)}
			{
				$F'(i,j) \gets F'(i,j) * 2y(i,j)$\;
				$y(i,j) \gets y(i,j)^2$\;
			}
			\;
			// Attenuating neurons\;
			\If{(neuron$(i,j)$ is the attenuating neuron)}
			{
				$F'(i,j) \gets F'(i,j) * (b-y(i,j)^2)/(y(i,j)^2+b)^2$\;
				$y(i,j)  \gets y(i,j)/(y(i,j)^2+b)$\;
			}
		}
	}
  \caption{The forward propagation of the neural network that contains the amplifying and attenuating neurons.}
  \label{al:forward}
\end{algorithm}
\begin{figure}
	\centering
	\begin{subfigure}{0.8\linewidth}
		\centering
		\includegraphics[width=\linewidth]{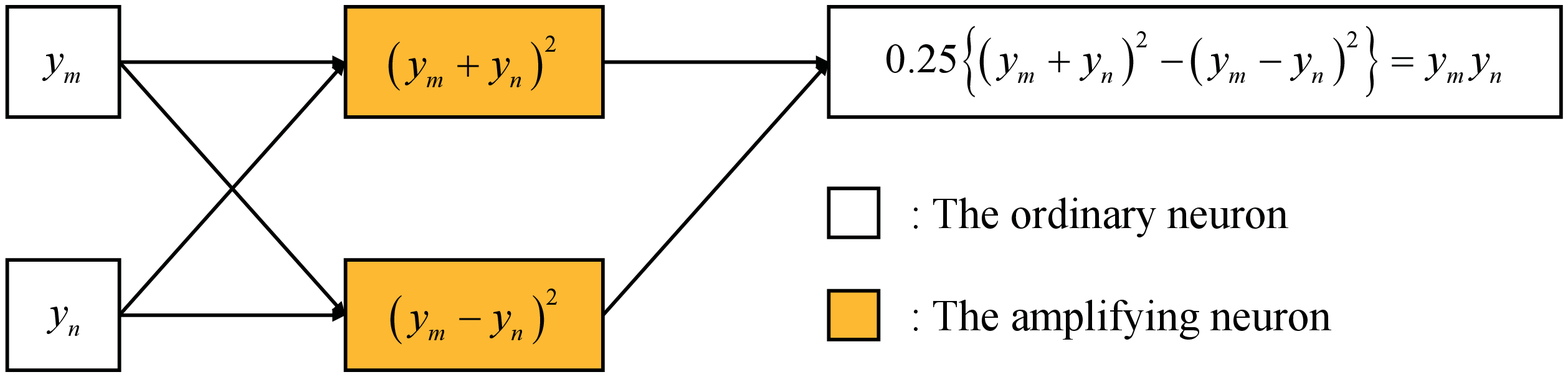}
		\caption{The multiplication operation performed in the hidden layers}
	\end{subfigure}
	\begin{subfigure}{0.8\linewidth}
		\centering
		\includegraphics[width=\linewidth]{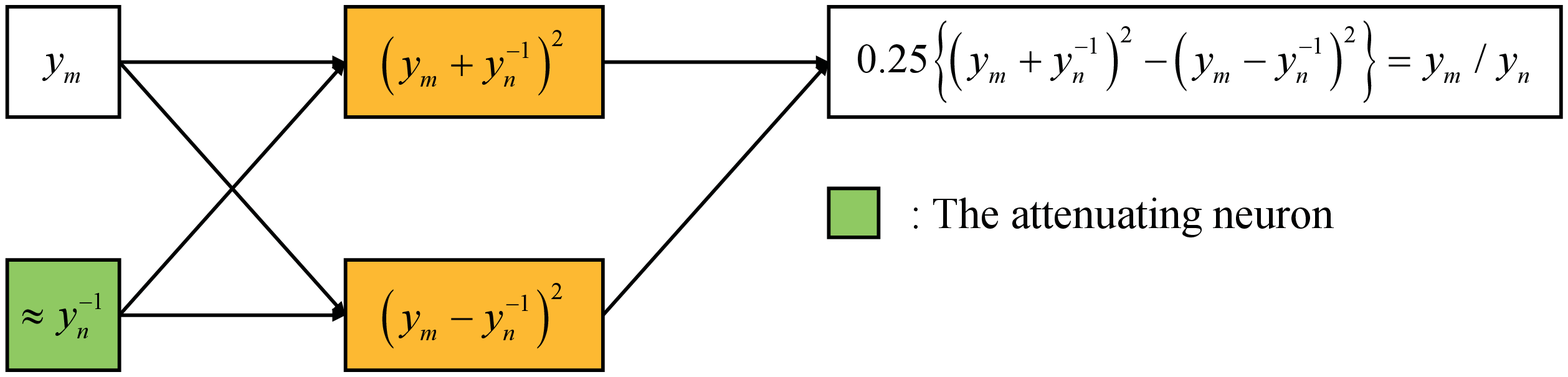}
		\caption{The division operation performed in the hidden layers}
	\end{subfigure}
	\caption{The representative neural network structures that can perform multiplication and division in the hidden layers.}
	\label{fig:multiplication_division}
\end{figure}
Using the function composition, the amplifying and attenuating neurons can be easily implemented into the neural network, as described in algorithm \ref{al:forward}, without any significant additional computational effort.
For the amplifying and attenuating neurons, the $H(x)$ value is no longer needed once the secondary activation function $G$ is applied.
Therefore, no additional memory is required for the amplifying and attenuating neurons.
For the error back-propagation, the activation function derivative of the amplifying and attenuating neurons can be calculated during the forward propagation process by using the chain rule.
Once the activation function derivative is properly calculated, the error back-propagation method can be used without any modification.

With the amplifying and attenuating neurons, the performance of the neural network can be improved.
For instance, the order of the neural network that contains the amplifying neuron increases, as the high-order activation function is provided.
In addition, theoretically, the neural network with the amplifying and attenuating neurons can perform multiplication and approximated division, as depicted in figure \ref{fig:multiplication_division}.
It is expected that such characteristics enhance the efficiency of the neural network;
i.e., the neural network with the amplifying and attenuating neurons can handle more complicated data without increasing the network size.

However, high-order models are more unstable in general, compared to first-order models.
Therefore, replacing the entire neurons of the neural network with the amplifying and attenuating neurons is likely to cause numerical problems, especially for the large neural network.
For instance, the order of the neural network only having high-order activation functions increases without limits, as the number of stacked hidden layers increases.
Such a numerical instability problem can be resolved by limiting the number of the amplifying and attenuating neurons, provided that the rest of neurons have first-order activation functions.

\section{Experiments}

To evaluate the performance improvement of the neural network with the amplifying and attenuating neurons, one-dimensional and two-dimensional problems were considered.
In the performance assessments, the multilayer perceptron model was adopted as a representative neural network model.
\begin{equation} \label{eq:psoftplus}
	f(x) = ax + (1-a) \ln (e^x + 1) = x + (1-a) \ln (e^{-x} + 1)
\end{equation}
Here, the parametric softplus function, defined in Eq (\ref{eq:psoftplus}), with $a$ of 0.3 was used as the activation function $F(x)$ for the ordinary neuron, and also as the first activation function $H(x)$ for the amplifying and attenuating neurons.
For the bias neuron and the neuron in the output layer, no activation function was applied.

The $L_2$ weight regularization method was used to minimize overfitting phenomenon.
Also, for the robust training of the neural network, ADAM method \citep{kingma2014adam} was utilized.
In the training process, a batch size of one was used, since the neural network trained with small batch methods generally outperforms the one with large batch methods \citep{keskar2016large}.
In the present study, for both numerical experiments, the best evaluation out of ten independent evaluations was dealt for each of network configurations.
All the training data sets, trained networks, and source code dealt are available at
\begin{center}
  \url{https://github.com/TheWinterSky/Amplifying-and-attenuating-neurons}
\end{center}

\subsection{One-dimensional problem}

\begin{figure}[ht]
	\centering
	\begin{subfigure}{0.45\linewidth}
		\centering
		\includegraphics[width=\linewidth]{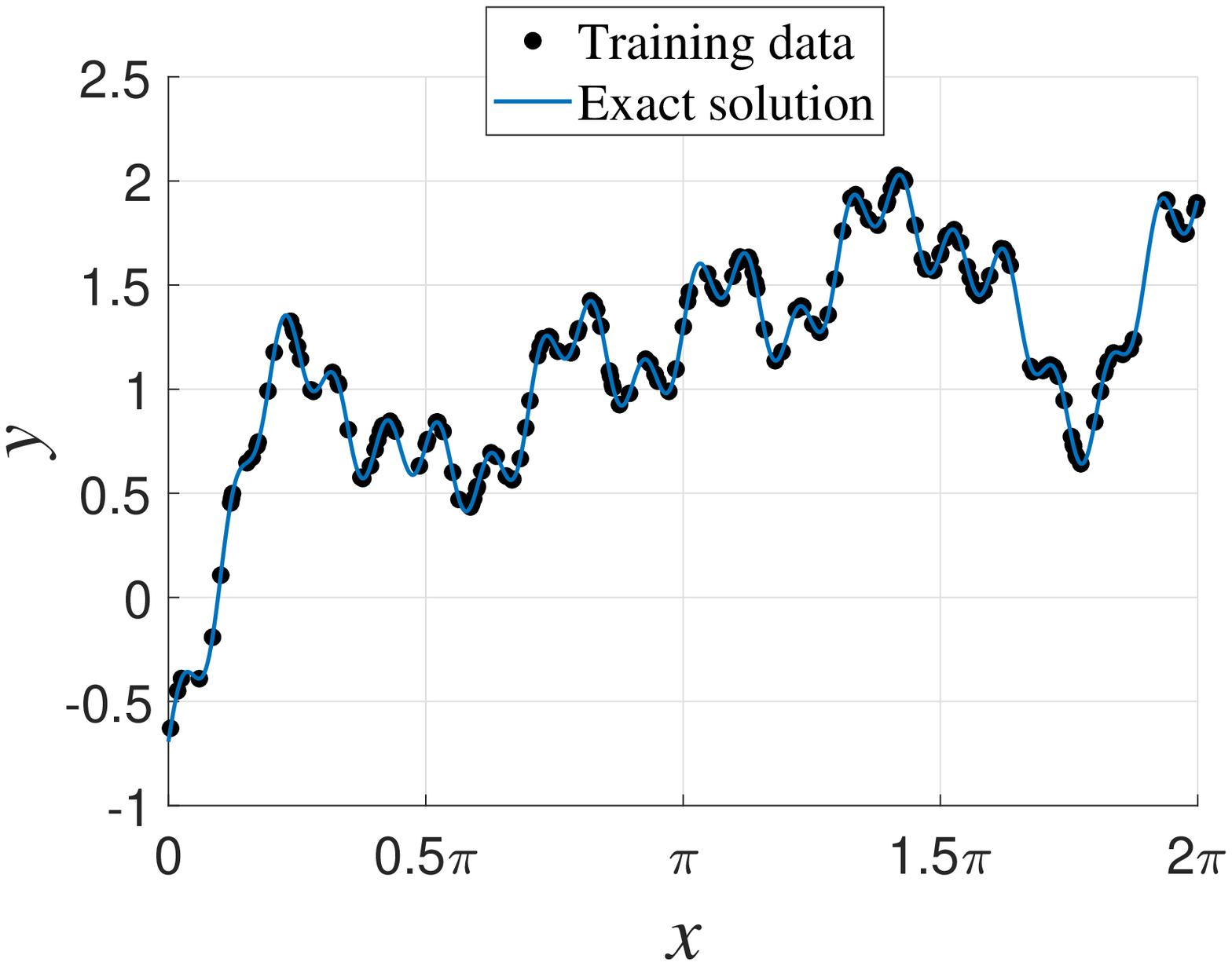}
		\caption{The exact solution and training data set that consists of 194 points}
	\end{subfigure}
	\begin{subfigure}{0.45\linewidth}
		\centering
		\includegraphics[width=\linewidth]{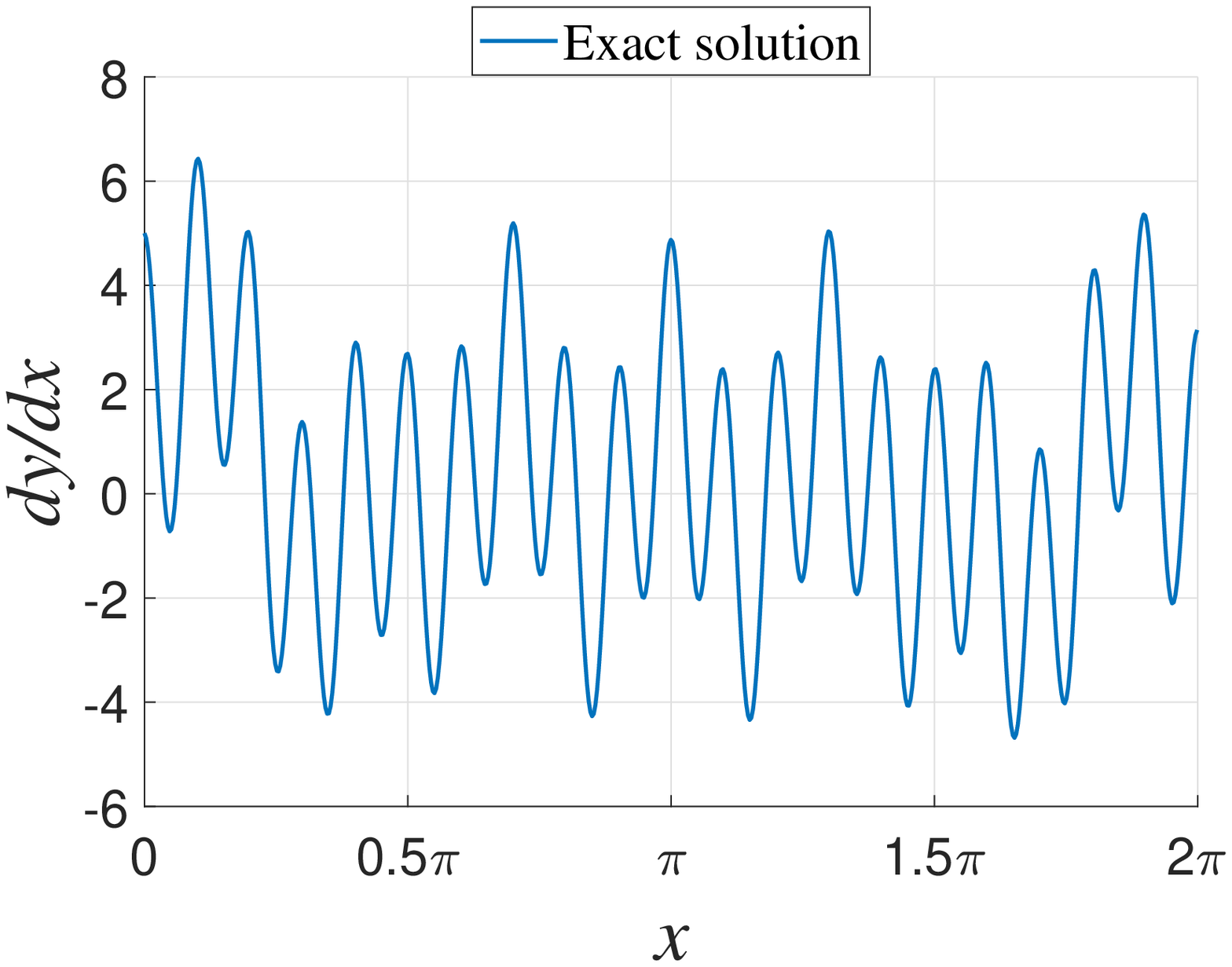}
		\caption{The derivative of the exact solution}
	\end{subfigure}
	\caption{The exact solution behavior and training data set distribution.}
	\label{fig:1D_exact_f}
\end{figure}
In the first numerical experiment, neural networks were required to reproduce a one-dimensional function.
The one-dimensional function (i.e., the exact solution) consists of multiple transcendental functions, as shown in Eq. (\ref{eq:1D_exact_f}), so that it cannot be exactly expressed by the finite Taylor series.
Such a characteristic is necessary for evaluating the performance improvement of the high-order neural network model, since the output of the neural network with the amplifying and attenuating neurons may possess a form of polynomials.
As shown in figure \ref{fig:1D_exact_f}, the exact solution shows a clear tendency over the interval $[0,2\pi]$, and exhibits an alternating sign of its derivative.
Therefore, the neural networks were required to reproduce not only the overall tendency, but also the high frequency feature of the exact solution.
To construct the training data set, 194 discrete pairs of $x$ and their corresponding exact solution values were used.
The network configurations and evaluation results (i.e., the mean absolute error (MAE) and standard deviation (SD) of error) are summarized in table \ref{table:1D}.
\begin{equation} \label{eq:1D_exact_f}
	y=\ln (x+0.5) + 0.2 \sin{x} + 0.4 \sin{2x} + 0.3 \sin{3x} - 0.1 \sin{5x} - 0.2 \sin{7x} + 0.15 \sin{20x}
\end{equation}
\begin{table}
	\caption{The MAE and SD of the error from the neural networks according to the network configurations.}
	\label{table:1D}
	\centering
	\begin{tabular}{lcccccc}
		\toprule
		& & \multicolumn{3}{c}{Per single hidden layer} & \multicolumn{2}{c}{Error ($x \in [0, 2\pi]$)} \\
		\cmidrule(r){3-5}
		\cmidrule(r){6-7}
		Network 			& Depth 				& 
		$n_\text{total neurons}$	& $n_\text{amplifying}$ & $n_\text{attenuating}$ & 
		MAE		& SD \\
		\midrule
		Network 1 & 5 & 10 & 0 & 0 & 0.085079 & 0.102091 \\
		Network 2 & 5 & 10 & 1 & 0 & 0.048816 & 0.066942 \\
		Network 3 & 5 & 10 & 0 & 1 & 0.072480 & 0.090168 \\
		Network 4 & 5 & 10 & 5 & 0 & 0.005758 & 0.013647 \\
		Network 5 & 5 & 10 & 4 & 1 & 0.005485 & 0.008369 \\
		Network 6 & 5 & 10 & 0 & 5 & 0.013222 & 0.023051 \\
		Network 7 & 9 & 10 & 0 & 0 & 0.003283 & 0.006276 \\
		Network 8 & 9 & 10 & 4 & 1 & 0.002212 & 0.005303 \\
		\bottomrule
	\end{tabular}
\end{table}
\begin{figure}
	\centering
	\begin{subfigure}{0.45\linewidth}
		\centering
		\includegraphics[width=\linewidth]{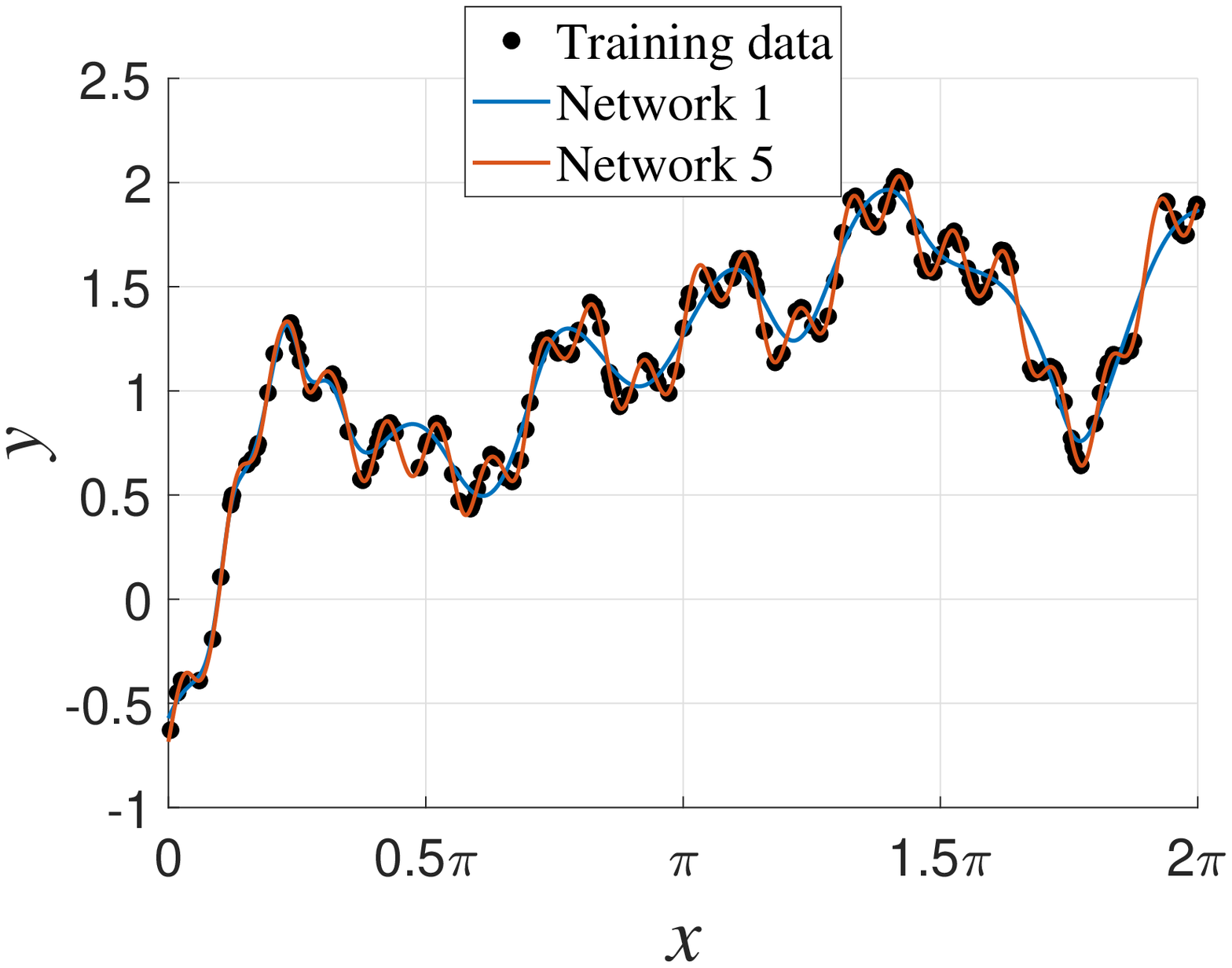}
		\caption{The training data set and results}
	\end{subfigure}
	\begin{subfigure}{0.45\linewidth}
		\centering
		\includegraphics[width=\linewidth]{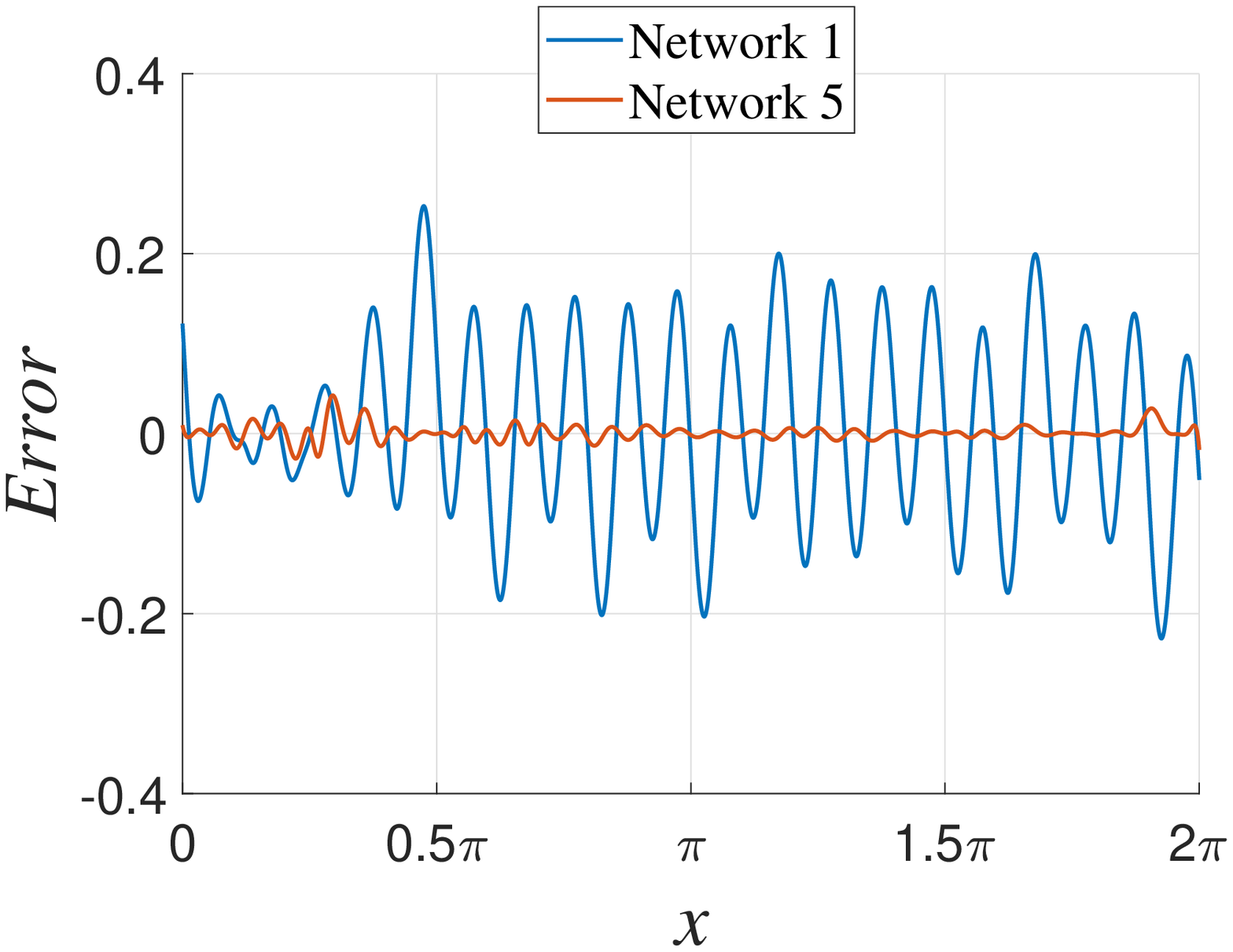}
		\caption{The error}
	\end{subfigure}
	\caption{The results $y$ and error (i.e., $y-y_\text{exact}$) from Network 1 and 5.}
	\label{fig:1D_Result}
\end{figure}
To evaluate the effect of the amplifying and attenuating neurons, the MAE and SD of error were compared for the neural networks of an identical size, i.e., Network 1\textasciitilde{}6.
From Network 2 to 6, the amplifying and attenuating neurons are placed to the entire hidden layers.
As shown in figure \ref{fig:1D_Result}, Network 1, which does not have any amplifying and attenuating neurons, successfully predicts the overall tendency of the exact solution, although the detailed behavior, such as the high frequency feature, is not reproduced properly because of the insufficient network size.
It was found that the performance of the neural networks increases when the amplifying and attenuating neurons are placed in the networks.
In addition, the neural networks with the amplifying neurons produce better results, compared to those with the attenuating neurons.
Nonetheless, Network 5 that contains both the amplifying and attenuating neurons, not Network 6, produces the most accurate results among the networks of an identical size, and Network 5 successfully reproduces the high frequency feature of the exact solution as shown in the figure \ref{fig:1D_Result}.
It is estimated that Network 5 outperforms Network 4 because reciprocal features are provided by the attenuating neuron for Network 5.

It is usually known that the main concern of large networks is the occurrence of overfitting phenomenon.
To examine whether the neural network with the amplifying and attenuating neurons suffers from overfitting, the neural networks of a sufficiently large size were also considered.
As observed from the MAE of Network 7, the neural network with nine hidden layers is enough to handle the exact solution.
In order to avoid the order of Network 8 being excessively high, the amplifying and attenuating neurons were placed from the first hidden layer to the fifth.
Compared to Network 7, no significant degradation of the output was observed for Network 8 in the continuous domain;
therefore, overfitting phenomenon did not occur.

\subsection{Two-dimensional problem}

\begin{figure}
	\centering
	\begin{subfigure}{0.45\linewidth}
		\centering
		\includegraphics[width=\linewidth]{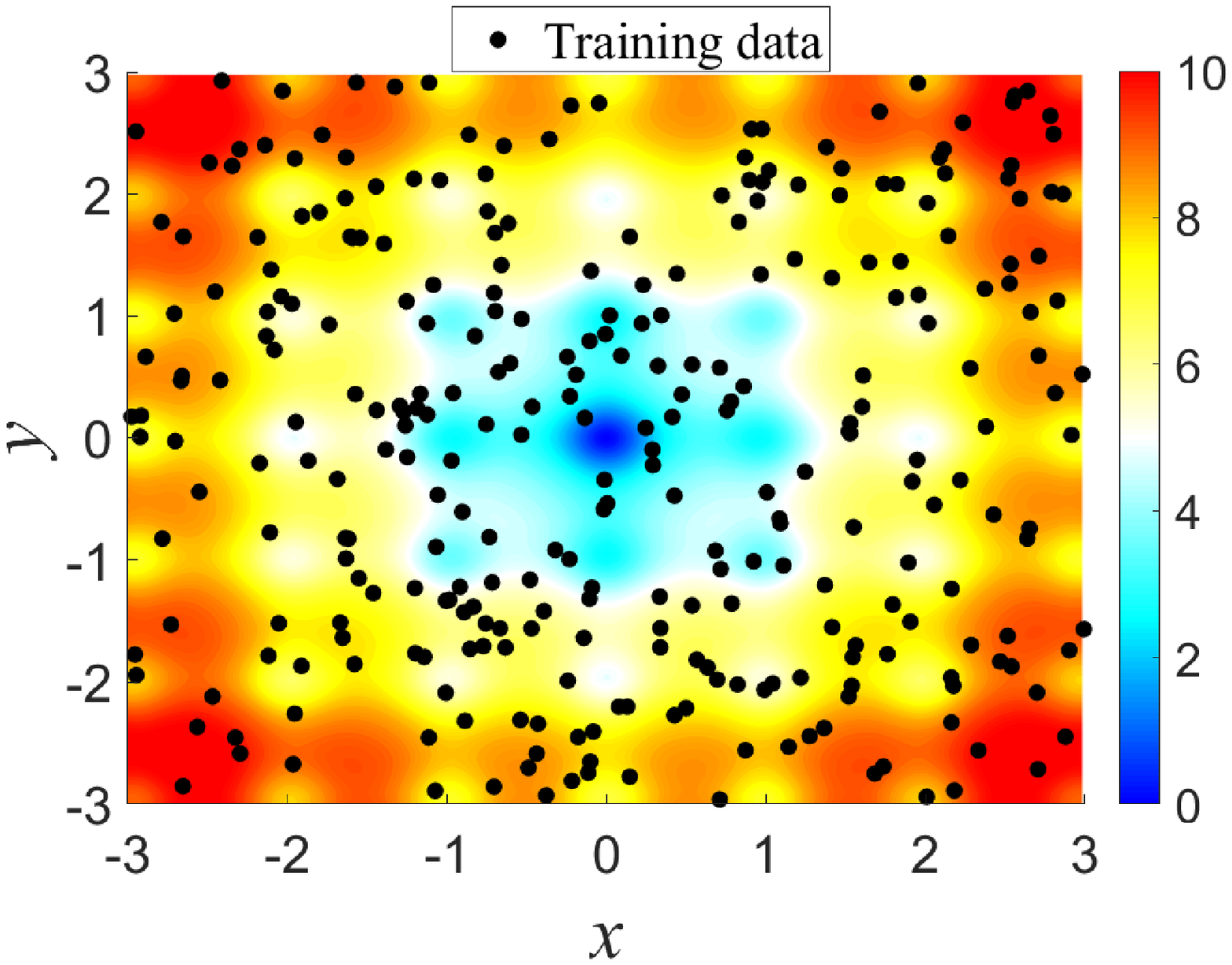}
		\caption{The contours of the exact solution and training data set that consists of 300 points}
	\end{subfigure}
	\begin{subfigure}{0.45\linewidth}
		\centering
		\includegraphics[width=\linewidth]{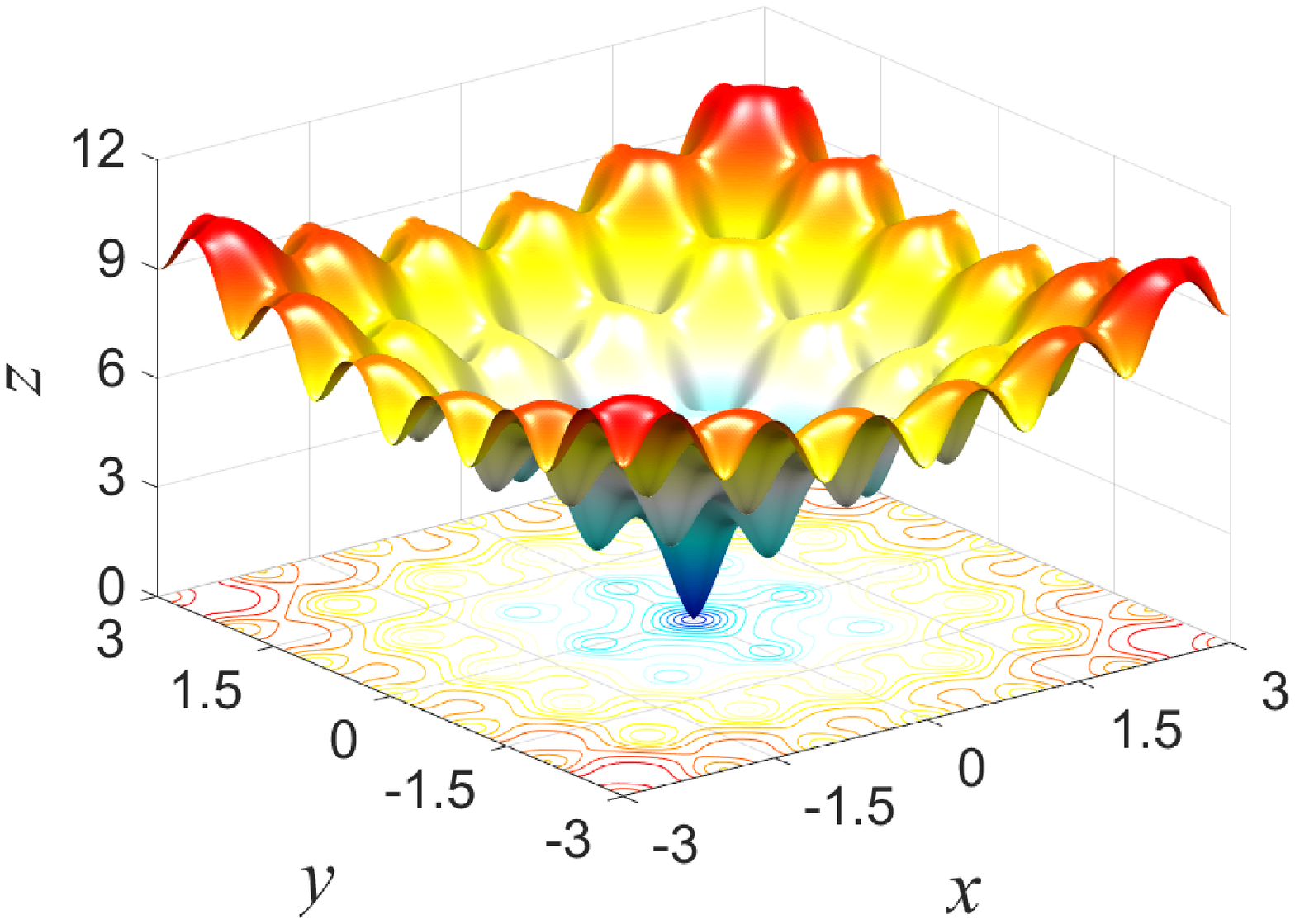}
		\caption{The exact solution in three-dimensional space}
	\end{subfigure}
	\caption{The behavior of the exact solution and training data set}
	\label{fig:2D_exact_f}
\end{figure}
\begin{figure}
	\centering
	\begin{subfigure}{0.45\linewidth}
		\centering
		\includegraphics[width=\linewidth]{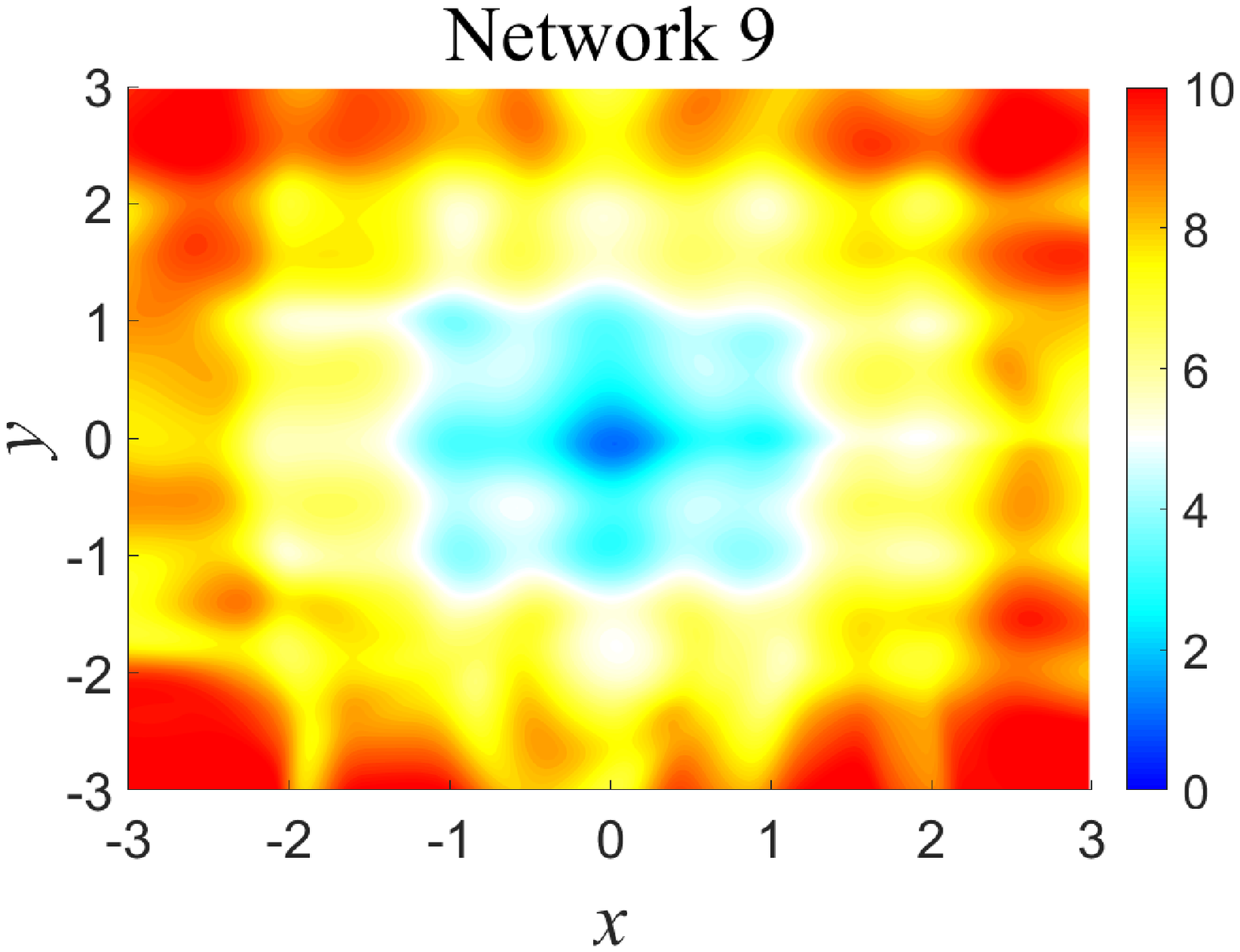}
		\caption{The contours from Network 9}
	\end{subfigure}
	\begin{subfigure}{0.45\linewidth}
		\centering
		\includegraphics[width=\linewidth]{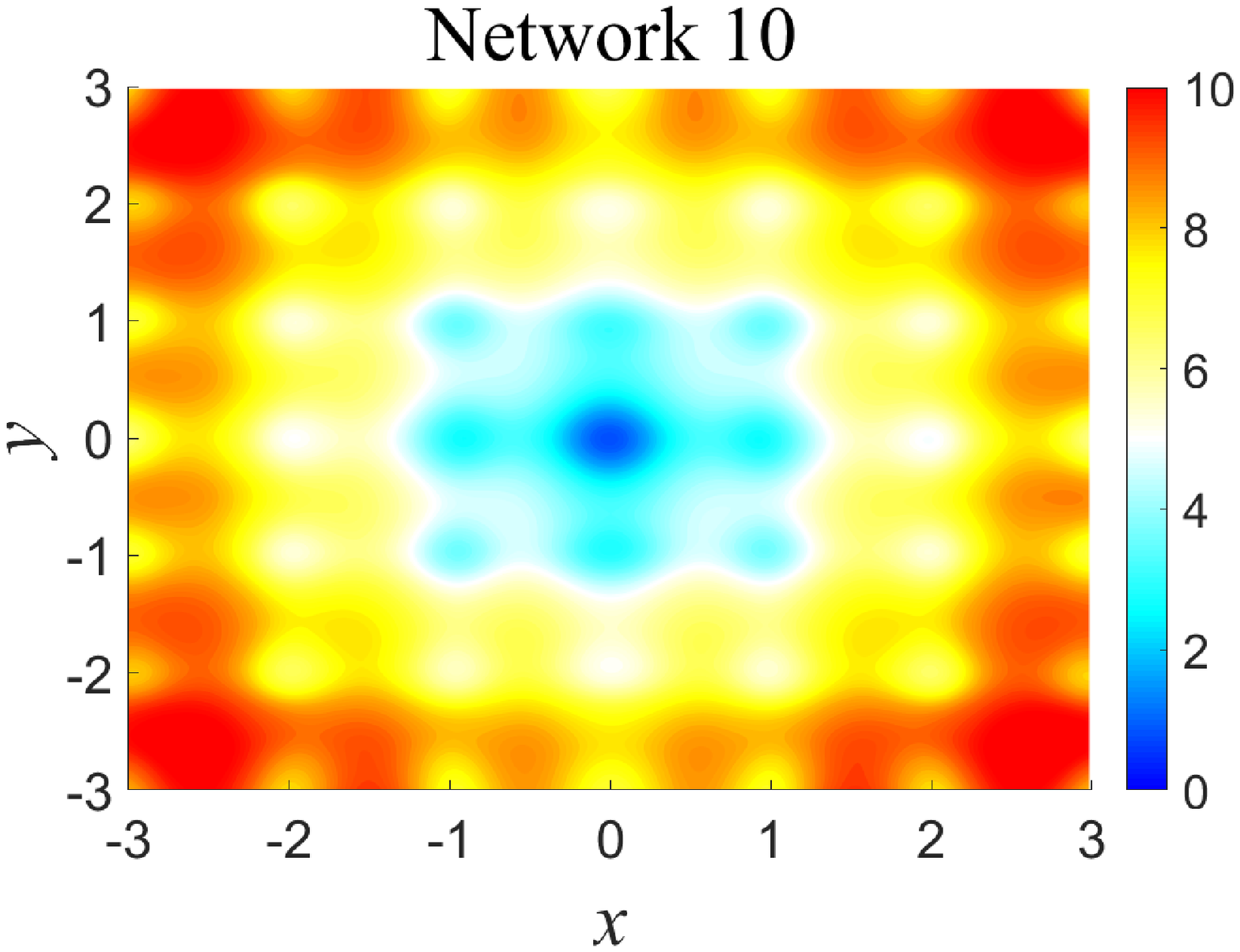}
		\caption{The contours from Network 10}
	\end{subfigure}
	\begin{subfigure}{0.45\linewidth}
		\centering
		\includegraphics[width=\linewidth]{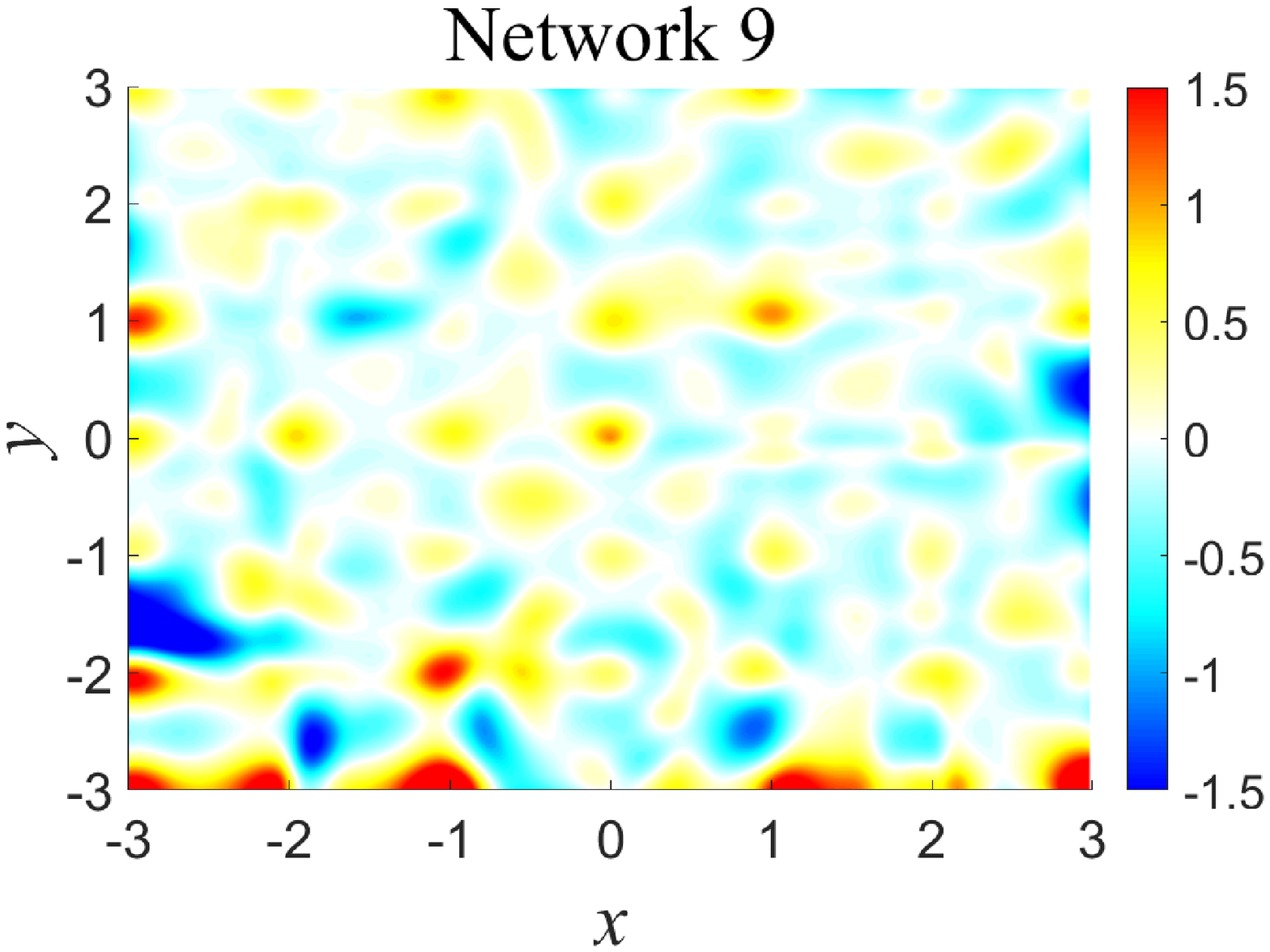}
		\caption{The error from Network 9}
	\end{subfigure}
	\begin{subfigure}{0.45\linewidth}
		\centering
		\includegraphics[width=\linewidth]{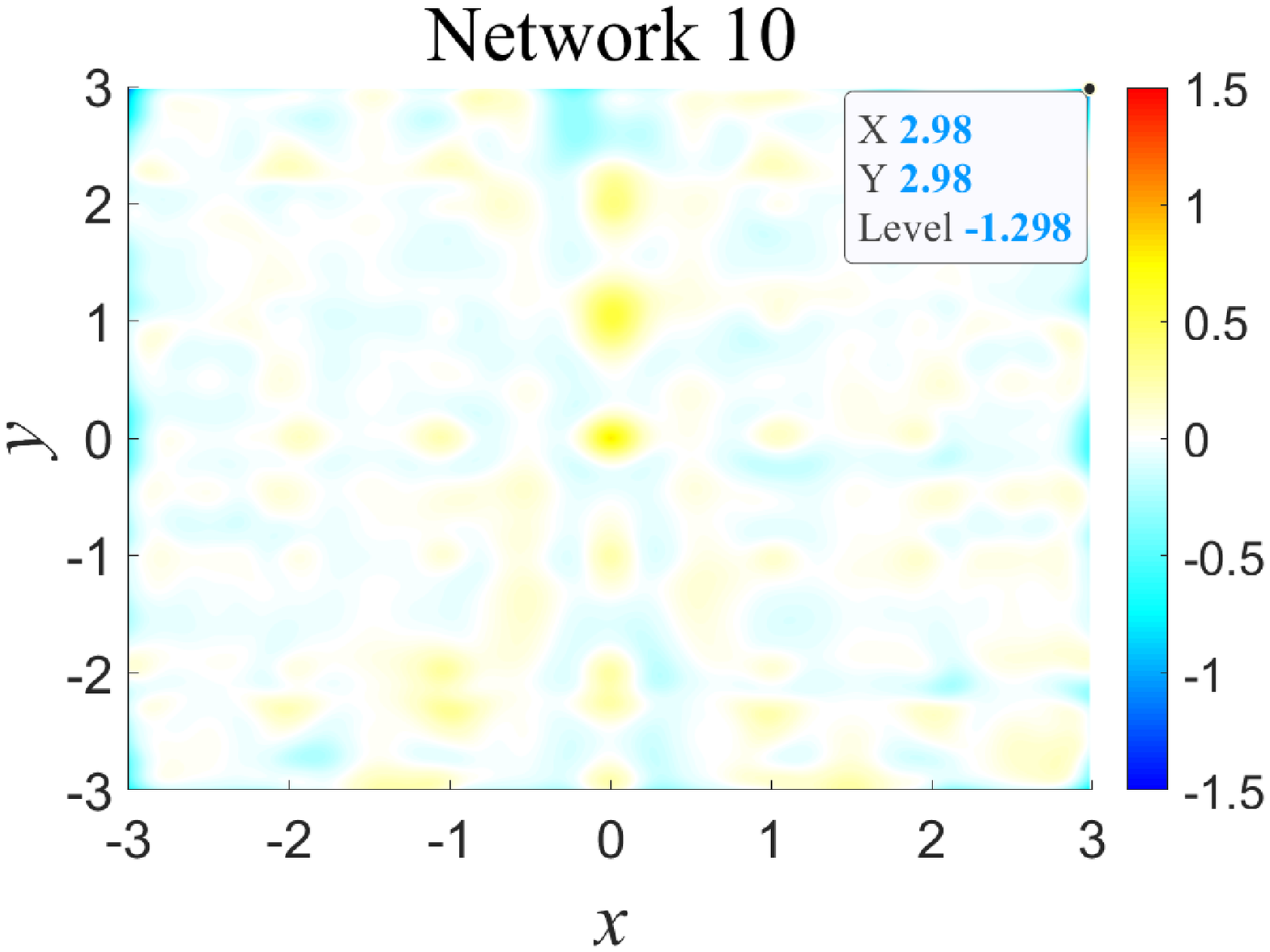}
		\caption{The error from Network 10}
	\end{subfigure}
	\caption{The output $y$ and error (i.e., $y-y_\text{exact}$) from Network 9 and 10.}
	\label{fig:2D_Result}
\end{figure}

As the second numerical experiment, the two-dimensional Ackley function \citep{ackley1987connectionist} was chosen to be the exact solution.
The Ackley function consists of exponential and sinusoidal functions as in Eq. (\ref{eq:2D_exact_f}).
It was required that neural networks reproduce the exact solution based on the training data set that consists of 300 randomly chosen discrete points illustrated in figure \ref{fig:2D_exact_f}.
The MAE and SD of the error from the evaluations and the network configurations are shown in table \ref{table:2D}.
To limit the order of Network 10, the amplifying and attenuating neurons are placed from the second hidden layer to the fifth for Network 10.
\begin{equation} \label{eq:2D_exact_f}
	f(x,y)=-20 \exp{\left( -0.2 \sqrt{\cfrac{x^2+y^2}{2}} \right)}
	       -   \exp{\left( \cfrac{\cos{2 \pi x} + \cos{2 \pi y}}{2} \right)}
	       + 20 + \exp{(1)}
\end{equation}
From the comparison of the MAE and SD of error between Network 9 and 10, it was observed that the absolute error drastically reduces when the amplifying and attenuating neurons are placed in the neural network as shown in figure \ref{fig:2D_Result}.
Both the MAE and SD of error decrease over the entire continuous domain area, especially in the region included in the training data set.
However, it was found that the maximum value of the absolute error was developed from outside of the training data set.
Such a behavior is considered to originate from the nature of the high-order model.
Generally, high-order models are not appropriate for extrapolation, although they outperform the first-order models when interpolation is performed instead.
\begin{table}
	\caption{The neural network performance comparisons by network configurations.}
	\label{table:2D}
	\centering
	\begin{tabular}{lcccccc}
		\toprule
		& & \multicolumn{3}{c}{Per single hidden layer} & \multicolumn{2}{c}{Error ($x,y \in [-3, 3]$)} \\
		\cmidrule(r){3-5}
		\cmidrule(r){6-7}
		Network 			& Depth 				& 
		$n_\text{total neurons}$	& $n_\text{amplifying}$ & $n_\text{attenuating}$ & 
		MAE		& SD \\
		\midrule
		Network 9  & 6 & 10 & 0 & 0 & 0.238485 & 0.382490 \\
		Network 10 & 6 & 10 & 3 & 1 & 0.056000 & 0.089219 \\
		\bottomrule
	\end{tabular}
\end{table}

Such a numerical behavior in the region outside of the training data set can be alleviated when the input of the amplifying neuron is normalized.
Hence, Batch Normalization method is expected to relieve the problem, since all the input distributions of the hidden neurons are rescaled and shifted.
Nonetheless, Batch Normalization also has a possibility of limiting the performance improvement of the neural network that contains the amplifying and attenuating neurons, since Batch Normalization can suppress the high-order characteristics itself.

\section{Conclusion}

In the present study, an amplifying neuron and attenuating neuron are proposed for improving the performance of the neural network.
Amplifying neurons square the activated output value of the neuron, and thus they can be considered as ordinary neurons with a second-order activation function.
Meanwhile, attenuating neurons yield an approximated reciprocal activated output value of neurons.
In theory, the order of the neural network increases when the amplifying neuron is placed.
Therefore, the neural network model with the amplifying neuron becomes a high-order model.
Furthermore, additional features can also be provided when the attenuating neuron is placed in the hidden layers.

From the numerical experiments, it was revealed that the neural networks that contain the amplifying and attenuating neurons outperform those without them.
In addition, the neural networks having both the amplifying and attenuating neurons yield the most accurate results.
Despite of the performance improvement, it was found that the output value of the neural networks containing the amplifying and attenuating neurons can be immoderate for the extrapolation region, outside of the training data set.
It is expected that such a behavior can be alleviated when Batch Normalization method is adopted, although it may limit the performance improvement.

\newpage
%
%
\bibliographystyle{unsrtnat}
\bibliography{References}

\end{document}